\documentclass[journal]{IEEEtran}
\usepackage{textcomp}
\usepackage{cite, amssymb}
\usepackage{makecell}
\usepackage{color, array,multirow}
\usepackage{algpseudocode}% http://ctan.org/pkg/algorithmicx
\usepackage{algorithm, amsmath}% http://ctan.org/pkg/algorithm
\usepackage{footnote}
\makesavenoteenv{tabular}
\makesavenoteenv{table}
\usepackage{multicol, blindtext}
\usepackage{dblfloatfix}

\ifCLASSINFOpdf
   \usepackage[pdftex]{graphicx}
\else
\fi
\begin{document}
\title{DropConnect Is Effective in Modeling Uncertainty of Bayesian Deep Networks}

\author{Aryan~Mobiny,~\IEEEmembership{Member,~IEEE,}
       Hien~V.~Nguyen,~\IEEEmembership{Member,~IEEE,}
       Supratik Moulik, Naveen Garg, Carol C. Wu
\thanks{A. Mobiny and H. V. Nguyen are with the Department of Electrical and Computer Engineering, University of Houston, Houston,
TX, 77004 USA (e-mail: amobiny@uh.edu and hienvnguyen@uh.edu).}

}        
\maketitle

\begin{abstract}
Deep neural networks (DNNs) have achieved state-of-the-art performances in many important domains, including medical diagnosis, security, and autonomous driving. In these domains where safety is highly critical, an erroneous decision can result in serious consequences. While a perfect prediction accuracy is not always achievable, recent work on Bayesian deep networks shows that it is possible to know when DNNs are more likely to make mistakes. Knowing what DNNs do not know is desirable to increase the safety of deep learning technology in sensitive applications. Bayesian neural networks attempt to address this challenge. However, traditional approaches are computationally intractable and do not scale well to large, complex neural network architectures. In this paper, we develop a theoretical framework to approximate Bayesian inference for DNNs by imposing a Bernoulli distribution on the model weights. This method, called MC-DropConnect, gives us a tool to represent the model uncertainty with little change in the overall model structure or computational cost. We extensively validate the proposed algorithm on multiple network architectures and datasets for classification and semantic segmentation tasks. We also propose new metrics to quantify the uncertainty estimates. This enables an objective comparison between MC-DropConnect and prior approaches. Our empirical results demonstrate that the proposed framework yields significant improvement in both prediction accuracy and uncertainty estimation quality compared to the state of the art.

% gained massive attention and been revolutionary in applied machine learning fields, and computer vision in particular. However, these models do not yield a prediction uncertainty measure to explain whether the model is certain about its output or not. This information is of critical in scenarios where control is handed to automated systems, autonomous driving cars and medical diagnosis for example.  Bayesian neural networks (BNNs) are the probabilistic version of DNNs capable of handing uncertainty. However, such Bayesian models are computationally intractable and do not scale well to large, complex neural network architectures. 

\end{abstract}

% Note that keywords are not normally used for peer review papers.

% \fi
\IEEEpeerreviewmaketitle

\vspace{-4mm}
\section{Introduction}
\IEEEPARstart{D}{eep} neural networks have revolutionized various applied fields, including engineering and computer science (such as AI, language processing and computer vision) \cite{krizhevsky2012imagenet, he2017mask, mnih2013playing}, as well as classical sciences (biology, physics, medicine, etc.) \cite{anjos2015neural, mobiny2018fast}. DNNs can learn abstract concepts and extract desirable information from the high dimensional input. This is done through stacks of convolutions followed by appropriate non-linear rectifiers. DNNs alleviate the need for time-consuming hand-engineered algorithms. Due to the high model complexity, DNNs  require a huge amount of data to regularize the training and prevent the networks from overfitting the training examples. This reduces their applicability in settings where data are scarce. This is often the case in scenarios where data collection is expensive or time-consuming, e.g. annotation of computed tomography scans by radiologists.

More importantly, popular deep learning models are often trained with maximum likelihood (ML) or maximum a posteriori (MAP) procedures, thus produce a point estimate but not an uncertainty value. In a classifier model, for example, the probability vector obtained at the end of the pipeline (the softmax output) is often erroneously interpreted as model confidence. In reality, a model can be uncertain in its predictions even with a high softmax output. In other words, the softmax probability is the probability that an input is a given class relative to the other classes, thus it does not help explain the model’s overall confidence \cite{gal2016uncertainty}. 

In applications of the automated decision making or recommendation systems, which might involve life-threatening situations, information about the \textit{reliability} of automated decisions is crucial to improve the system's safety. In other words, we want to know how confident the model is about its predictions \cite{der2009aleatory}.
Understanding if the model is under-confident or falsely over-confident can inform users to perform necessary actions to ensure the safety \cite{gal2016uncertainty}. Take an automated cancer detection system as an example which might encounter an out-of-distribution test sample. A traditional DNN-based system makes unreasonable suggestions, and as a result, unjustifiably bias the expert. However, given information about the model's confidence, an expert could rely more on his own judgment when the automated system is essentially guessing at random.

Most of the studies on uncertainty estimation techniques are inspired by Bayesian statistics. Bayesian Neural Networks (BNNs) \cite{neal2012bayesian} are the probabilistic version of the traditional NNs with a prior distribution on its weights. Such networks are intrinsically suitable for generating uncertainty estimates as they produce a distribution over the output for a given input sample \cite{mackay1992practical}. However, such models are computationally expensive in practice because of the huge number of parameters in the large models commonly used  and the computationally intractable inference of the model posterior. Thus much effort has been spent on developing scalable, approximated BNNs.

Variational inference \cite{blei2017variational} is the most common approach used for approximating the model posterior using a simple variational distribution such as Gaussian \cite{graves2011practical}. The parameters of the distribution are then set in a way that it is as similar as possible to the true distribution (usually by minimizing the Kullback-Leibler divergence). However, the use of the Gaussian distribution considerably increases the required number of parameters and makes it computationally expensive. 

In this paper, we propose a mathematically-grounded method called Monte Carlo DropConnect (MC-DropConnect) to approximate variational inference in BNNs. The main contributions of this paper are:

\begin{enumerate}
\item We propose imposing the Bernoulli distribution \textit{directly} to the weights of the deep neural network to estimate the posterior distribution over its weight matrices. We derive the required equations to show that this generalization provides a computationally tractable approximation of a BNN, only using the existing tools and no additional model parameters or increase in time complexity.

\item We extensively evaluate the proposed MC-DropConnect method in image classification and semantic segmentation tasks. Our experimental results show the robust generalization of MC-DropConnect, and considerable improvement in the prediction accuracy compared to standard techniques.

\item We propose metrics to evaluate the uncertainty estimation performance of the Bayesian models in the classification and segmentation settings. Using these metrics, we show that our method is superior compared to the recently proposed technique called MC-Dropout. 

% the uncertainty estimation of our proposed method with that of a recently proposed technique called MC-Dropout, and show our superior precision .

\item We make an in-depth analysis of the uncertainty estimations in both classification and segmentation settings and show that uncertainty informed decisions can be used for improving the performance of deep networks. 

\end{enumerate}

Our experimental results (achieved using the proposed method and metrics) provide a new benchmark for other researchers to evaluate and compare their uncertainty estimation in pursuit of safer and more reliable deep networks. The rest of this paper is organized as follows: works related to approximating Bayesian inference and estimating uncertainty are presented in Section \ref{related_work}. Section~\ref{methodology} explains our method proposed to approximate variational inference in Bayesian networks. Experimental results are presented and discussed in Section~\ref{results}. Section~\ref{conclusion} concludes the paper with future research directions.

%   This is different from MC-Dropout technique which applies the Bernoulli distribution to the network activation units, i.e. defining a single Bernoulli distribution over a group of weights exiting from a single neuron.   

\section{Related work}
\label{related_work}

Many studies have been conducted on approximate Bayesian inference for neural networks using deterministic approaches \cite{mackay1992practical}, Markov Chain Monte Carlo with Hamiltonian Dynamics \cite{neal1993bayesian} and variational inference \cite{graves2011practical}. In particular, \cite{neal1993bayesian} introduced the Hamiltonian Monte Carlo for Bayesian neural network learning which gives a set of posterior samples. This method does not require the direct calculation of the posterior but is computationally prohibitive. 

Recently, Gal et al. \cite{gal2016dropout} showed that Dropout, a well-known regularization technique \cite{srivastava2014dropout}, is mathematically equivalent to approximate variational inference in the deep Gaussian process \cite{damianou2013deep}. This method, commonly known as MC-Dropout, uses Bernoulli approximating variational distribution on the network units and introduces no additional parameters for the approximate posterior. The main disadvantage of this method is that it often requires many forward-pass sampling which makes it resource-intensive\cite{gal2015bayesian}. Moreover, a fully Bayesian network approximated using this method (i.e. dropout applied to all layers) results in a too strong regularizer \cite{kendall2015bayesian} that learns slowly and does not achieve high prediction accuracy. Multiplicative Normalizing Flows \cite{louizos2017multiplicative} is another technique which is introduced as a family of approximate posteriors for the parameters of a variational BNN, capable of producing uncertainty estimates. However, it does not scale to very large convolutional networks. 

Another proposed approach is  Deep Ensembles \cite{lakshminarayanan2017simple} which shown to be able to achieve superior high-quality uncertainty estimates. This method takes the frequentist approach to estimate the model uncertainty by training several models and calculating the variance of their output prediction. However, this technique is quite resource-intensive as it requires to store several separate models and perform forward passes through all of them to make the inference. An alternative to such methods is \cite{devries2018learning} which proposes to \textit{learn} uncertainty from the given input.

\section{Methodology}
\label{methodology}

In this section, we briefly review Bayesian Neural Network and its limitations, variational inference as the standard technique in Bayesian modeling, and DropConnect, a method for regularization in NNs. We then use these tools to approximate a Bayesian framework with a standard neural network equipped with Bernoulli distribution applied to all its weights \textit{directly}. Finally, we explain the methods used for measuring and evaluating the model uncertainty.

\subsection{Bayesian Neural Networks}

From a probabilistic perspective, standard NN training via optimization is equivalent to maximum likelihood estimation (MLE) for the weights. Using MLE ignores any uncertainty that we may have in the proper weight values. Bayesian Neural Networks (BNN) are the extension over NNs to address this shortcoming by placing a prior distribution (often a Gaussian) over a NN's weight. This brings vital advantages like automatic model regularization and uncertainty estimates on predictions \cite{mackay1992practical, graves2011practical}. 

Given a BNN model with L layers parametrized by weights $\mathbf{w}=\{\mathbf{W}_i\}_{i=1}^L$ and a dataset $\mathcal{D}=(\mathbf{X, y})$, the Bayesian inference is to calculate the posterior distribution of the weights given the data, $p(\mathbf{w}|\mathcal{D})$. The predictive distribution of an unknown label $\mathbf{y}^*$ of a test input data $\mathbf{x}^*$ is given by:

\begin{equation}
\label{eq1}
\begin{aligned}
    p(\mathbf{y}^*|\mathbf{x}^*,\mathcal{D}) & =\mathbb{E}_{p(\mathbf{w}|\mathcal{D})}[p(\mathbf{y}^*|\mathbf{x}^*, \mathbf{w})]\\ & =\int p(\mathbf{y}^*|\mathbf{x}^*, \mathbf{w}) p(\mathbf{w}|\mathcal{D}) \, d\mathbf{w}
\end{aligned}
\end{equation}

\noindent
which shows that making a prediction about the unknown label is equivalent to using an ensemble of an uncountably infinite number of neural networks with various configuration of the weights. This is computationally intractable for neural networks with any size. The posterior distribution $p(\mathbf{w}|\mathcal{D})$ also cannot generally be evaluated analytically. Therefore, so much effort has been put into approximating BNNs to make them easier to train \cite{mackay1995probable, blundell2015weight}. 

\begin{figure*}[t]
\begin{center}
\includegraphics[width=\linewidth]{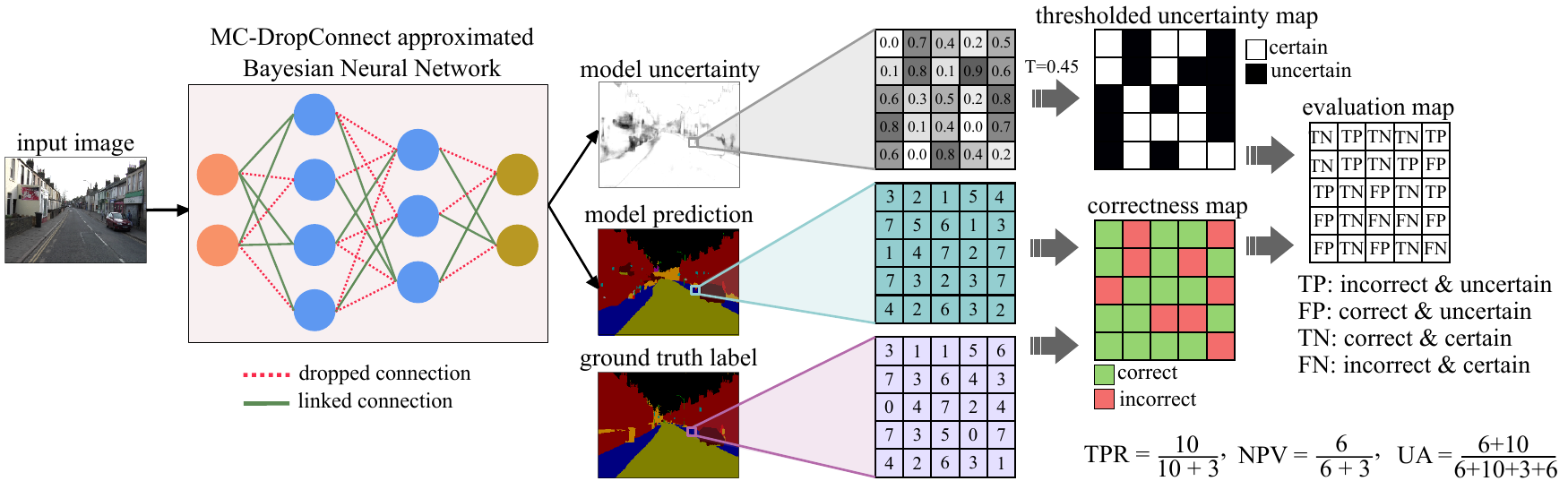}
\end{center}
   \caption{Overview of the proposed approximate Bayesian model (Left) and  metrics to evaluate the uncertainty quality (Right) in a semantic segmentation example. Since segmentation is identical to pixel-wise classification, similar computations hold true for the classification task.}
\label{fig:uncertainty_metrics}
\end{figure*}

One common approach is to use variational inference to approximate the posterior on the weights. It introduces a variational distribution on the weights, $q_\theta(\mathbf{w})$, parametrized on $\theta$ that minimizes the Kullback-Leibler (KL) divergence between $q$ and the true posterior:

\begin{equation}
\label{eq2}
\text{KL}(q_\theta(\mathbf{w}) || p(\mathbf{w}|\mathcal{D}))
\end{equation}

Minimising the KL divergence is equivalent to minimizing the negative evidence lower bound (ELBO):

\begin{equation}
\label{eq3}
\mathcal{L}(\theta) = -\int q_\theta(\mathbf{w}) \, \text{log}\,p(\mathbf{y}|\mathbf{X, w}) \,d\mathbf{w} +  \text{KL}(q_\theta(\mathbf{w}) || p(\mathbf{w}))
\end{equation}

\noindent
with respect to variational parameter $\theta$. The first term (commonly referred to as the \textit{expected log likelihood}) encourages $q_{\theta}(\mathbf{w})$ to place its mass on configurations of the latent variable that explains the observed data. However, the second term (referred to as \textit{prior KL}) encourages $q_\theta(\mathbf{w})$ to be similar to the prior, preventing the model from over-fitting. The prior KL term can be analytically evaluated for proper choice of prior and variational distribution. However,  the expectation (i.e. integral term) cannot be computed exactly for a non-linear neural network. Our goal in the next section is to develop an explicit and accurate approximation for this expectation. Our approach extends on the results of Gal et al. \cite{gal2015dropout} and uses Bernoulli approximating variational inference and Monte-Carlo sampling.

\subsection{DropConnect}
DropConnect \cite{wan2013regularization}, known as the generalized version of Dropout \cite{srivastava2014dropout}, is the method used for regularizing deep neural networks. Here, we briefly review Dropout and DropConnect applied to a single fully-connected layer of a standard NN. For a single $K_{i-1}$ dimensional input $\mathbf{v}$, the $i^{th}$ layer of an NN with $K_i$ units would output a $K_i$ dimensional activation vector $\mathbf{a}_i = \sigma(\mathbf{W}_i\mathbf{v})$ where $\mathbf{W}_i$ is the $K_i \times K_{i-1}$ weight matrix and $\sigma(.)$ the nonlinear activation function (biases included in the weight matrix with a corresponding fixed input of 1 for the ease of notation). 

    When Dropout is applied to the output of a layer, the output activations can be written as $\mathbf{a}_{i}^{\text{\tiny DO}} = \sigma(\mathbf{z}_i \odot (\mathbf{W}_i\mathbf{v}))$ where $\odot$ signifies the Hadamard product and $\mathbf{z}_i$ is a $K_i$ dimensional binary vector with its elements drawn independently from $z_{i}^{(k)}\sim\text{Bernoulli(}p_i\text{)}$ for $k=1, ..., K_i$ and $p_i$ to be the probability of keeping the output activation.

DropConnect, however, is the generalization of Dropout where the Bernoulli dropping is applied directly to each weight, rather than each output unit, thus the output activation is re-written as $\mathbf{a}_{i}^{\text{\tiny DC}} = \sigma((\mathbf{Z}_i \odot \mathbf{W}_i) \mathbf{v})$. Here, $\mathbf{Z}_i$ is the binary matrix of the same shape as $\mathbf{W}_i$, i.e. $K_i \times K_{i-1}$. Wan et al. \cite{wan2013regularization} showed that adding DroConnect helps regularize large neural network models and outperforms Dropout on a range of datasets.

\subsection{DropConnect for Bayesian Neural Network Approximation}
Assume the same Bayesian NN  with L layers parametrized by weights $\mathbf{w}=\{\mathbf{W}_i\}_{i=1}^L$. We perform the variational learning by approximating the variational distribution $q(\mathbf{W}_i|\boldsymbol{\Theta}_i)$ for every layer $i$ as:

\begin{equation}
    \label{eq4}
    \mathbf{W}_i = \boldsymbol{\Theta}_i \, \odot \, \mathbf{Z}_i
\end{equation}

\noindent
where $\boldsymbol{\Theta}_i$ is the matrix of variational parameters to be optimised, and $ \mathbf{Z}_i$ the binary matrix whose elements are distributed as:

\begin{equation}
    z_{i}^{(l, k)} \sim \text{Bernoulli}(p_i) \;\;\;\text{for} \;\; i=1, ..., L
\end{equation}

\noindent
Here, $z_{i}^{(l, k)}$ is the random binary value associated with the weight connecting the $l^{th}$ unit of the $(i-1)^{th}$ layer to the $k^{th}$ unit of the $i^{th}$ layer. $p_i$ is the probability that the random variables $z_{i}^{(l, k)}$ take the value 1 (assuming the same probability for all the weights in a layer). Therefore, $z_{i}^{(l, k)}=0$ corresponds to the weight being dropped out.

We start with rewriting the first term as a sum over all samples. Then we use equation (\ref{eq4}) to re-parametrize the integrand in equation (\ref{eq3}) so that it only depends on the Bernoulli distribution instead of $\mathbf{w}$ directly. We estimate the intractable integral with Monte Carlo sampling over $\mathbf{w}$ with a single sample as:

\begin{equation}
\begin{aligned}
   -\int q_\theta(\mathbf{w}) \, &\text{log}\,p(\mathbf{y}|\mathbf{X, w}) \,d\mathbf{w} \\ &=\sum_{n=1}^{N}\int -q_\theta(\mathbf{w}) \, \text{log}\,p(\mathbf{y}_n|\mathbf{x}_n, \mathbf{w})\\ & =\frac{1}{N}\sum_{n=1}^{N} -\text{log}\,p(\mathbf{y}_n|\mathbf{x}_n, \hat{\mathbf{w}}_n)
\end{aligned}
\end{equation}

Note that $\hat{\mathbf{w}}_n$ is not maximum a posteriori estimate, but random variable realisations from the Bernoulli distribution, $\hat{\mathbf{w}}_n \sim q_\theta(\mathbf{w})$, which is identical to applying DropConnect to the weights of the network. The final sum of the log probabilities is the loss of the NN, thus we set:

\begin{equation}
    \text{I}_{\text{NN}}(\mathbf{y}_n, \hat{\mathbf{y}}(\mathbf{x}_n, \hat{\mathbf{w}}_n)) = -\text{log}\,p(\mathbf{y}_n|\mathbf{x}_n, \hat{\mathbf{w}}_n)
\end{equation}

\noindent
where $\hat{\mathbf{y}}(\mathbf{x}_n, \hat{\mathbf{w}}_n)$ is the random output of the BNN. $\text{I}_{\text{NN}}$ is defined according to the task with the sum of squared loss and softmax loss commonly selected for the regression and classification respectively.

The second term in equation (\ref{eq3}) can be approximated following \cite{gal2015dropout}. It is been shown that the KL term is equivalent to $\sum_{i=1}^{L} ||\Theta_i||_2^2$. Thus, the objective function can be re-written as:

\begin{equation}
    \hat{\mathcal{L}}_{\text{MC}}=\frac{1}{N}\sum_{n=1}^N \text{I}_{\text{NN}}(\mathbf{y}_n, \mathbf{\hat{y}}_n)+\lambda\sum_{i=1}^L ||\Theta_i||_2^2
\end{equation}

\noindent
which is an scaled unbiased estimator of equation (\ref{eq3}). More interestingly, it is identical to the objective function used in an standard neural network with L2 weight regularization and DropConnect applied to all the weights of the network. Therefore, training such a neural network with stochastic gradient descent has the same effect as minimizing the KL term in equation (\ref{eq2}). This scheme, similar to a BNN, result in a set of parameters that best explains the observed data while preventing over-fitting.

After training the NN with DropConnect and proper regularization, we follow equation (\ref{eq1}) to make inference. We replace the posterior $p(\mathbf{w}|\mathcal{D})$ with the approximate posterior distribution $q_\theta(\mathbf{w})$ and approximate the integral with Monte Carlo integration:

\begin{equation}
\label{eq9}
\begin{aligned}
    p(\mathbf{y}^*|\mathbf{x}^*,\mathcal{D}) & \approx \int p(\mathbf{y}^*|\mathbf{x}^*, \mathbf{w}) q_\theta(\mathbf{w}) \, d\mathbf{w} \\& \approx \frac{1}{T}\sum_{t=1}^T p(\mathbf{y}^*|\mathbf{x}^*, \hat{\mathbf{w}}_t)=p_{\text{\tiny{MC}}}(\mathbf{y^*}|\mathbf{x^*})
\end{aligned}
\end{equation}

\noindent
with $\hat{\mathbf{w}}_t \sim q_{\theta}(\mathbf{w})$. This means that at test time, unlike common practice, the DropConnect layers is kept active to keep the Bernoulli distribution over the network weights. Then each forward pass through the trained network generates a Monte Carlo sample from the posterior distribution. Several of such forward passes are needed to approximate the posterior distribution of softmax class probabilities. According to equation (\ref{eq9}), the mean of these samples can be interpreted as the network prediction. We call this approach MC DropConnect which is a generalization over the previous work referred to as MC Dropout \cite{gal2015dropout} and will show its superiority in terms of achieving higher prediction accuracy and more precise uncertainty estimation in different ML tasks.

\subsection{Measuring the Model Uncertainty}
Generally, there are two types of uncertainty in Bayesian modelling \cite{der2009aleatory}. Model uncertainty, also known as Epistemic uncertainty, measures what the model does not know due to the lack of training data.  This uncertainty captures our ignorance about which model generated our collected data, thus can be explained away given enough data \cite{gal2016uncertainty}. Aleatoric uncertainty, however, captures noise (such as motion or sensor noise) inherent in the data and cannot be reduced by collecting more data \cite{kendall2017uncertainties}.

Having computed the result of stochastic forward passes through the model, we can estimate the model confidence to its output. In the classification setting, several metrics are introduced to measure uncertainty. One straightforward approach used in \cite{kendall2015bayesian} is to take the \textit{variance} of the MC samples from the posterior distribution as the output model uncertainty for each class. Predictive entropy is also suggested in \cite{gal2016uncertainty} which captures both epistemic and aleatoric uncertainty. However, this is not a suitable choice as we are more interested in regions of the data space where the model is uncertain.

\begin{table*}[!b]
\centering
\caption{Test prediction error (\%) and uncertainty estimation performance of the LeNet and FCNet networks and their Bayesian estimates on the MNIST and CIFAR-10 datasets.}
\begin{tabular}{cl|cc|ccc|}
\cline{3-7}
\multirow{2}{*}{} & \multicolumn{1}{c|}{\multirow{2}{*}{}} & \multicolumn{2}{c|}{Prediction Error (\%)} & \multicolumn{3}{c|}{Uncertainty metrics AUC (\%)} \\ \cline{3-7} 
 & \multicolumn{1}{c|}{} & Standard & MC-sampling & TPR & NPV & UA \\ \hline
\multicolumn{1}{|c|}{\multirow{3}{*}{\begin{tabular}[c]{@{}c@{}}MNIST\\ (LeNet-5)\end{tabular}}} & None & 0.99 & - & - & - & - \\
\multicolumn{1}{|c|}{} & MC-Dropout & 0.75 & 0.77 & 31.24 & 98.77 & 97.48 \\
\multicolumn{1}{|c|}{} & MC-DropConnect & \textbf{0.70} & \textbf{0.57} & \textbf{41.67} & \textbf{99.57} & \textbf{98.87} \\ \hline
\multicolumn{1}{|c|}{\multirow{3}{*}{\begin{tabular}[c]{@{}c@{}}CIFAR-10\\ (FCNet)\end{tabular}}} & None & 12.00 & - & - & - & - \\
\multicolumn{1}{|c|}{} & MC-Dropout & \textbf{10.92} & 10.57 & 38.24 & 92.12 & 82.89 \\
\multicolumn{1}{|c|}{} & MC-DropConnect & 11.34 & \textbf{10.15} & \textbf{40.29} & \textbf{94.31} & \textbf{87.27} \\ \hline
\end{tabular}
\label{classification_result}
\end{table*}

To specifically measure the model uncertainty for a new test sample $\mathbf{x^*}$, we can see it as the amount of information we would gain about the model parameters if we were to receive the true label $\mathbf{y^*}$. Theoretically, if the model is well-established in a region, knowing the output label conveys little information. In contrast, knowing the label would be informative in regions of data space where the model is uncertain \cite{smith2018understanding}. Therefore, the mutual information (MI) between the true label and the model parameters is as:

\begin{equation}
    I(\mathbf{\mathbf{y^*, w| \mathbf{x}^*, \mathcal{D}}}) = H(\mathbf{y}^*|\mathbf{x}^*, \mathcal{D})-\mathbb{E}_{p(\mathbf{w}|\mathcal{D})}[p(\mathbf{y}^*|\mathbf{x}^*, \mathbf{w})]
\end{equation}

\noindent
where given the training dataset $\mathcal{D}$, $\mathbf{y}^*$, $I(\mathbf{\mathbf{y^*, w| \mathbf{x}^*, \mathcal{D}}})$ measures the amount of information we gain about the model parameters $\mathbf{w}$ by receiving a test input $\mathbf{x}^*$ and its corresponding true label, $\mathbf{y}^*$. This can be approximated using the Bayesian interpretation of DropConnect derived earlier. $H$ is the entropy,  commonly known as the \textit{predictive entropy}, which captures the amount of information in the predictive distribution:

\begin{equation}
    H(\mathbf{\mathbf{y^*| \mathbf{x}^*, \mathcal{D}}}) = - \sum_c p(\mathbf{y}^*=c|\mathbf{x}^*, \mathcal{D})\,\text{log} \, p(\mathbf{y}^*=c|\mathbf{x}^*, \mathcal{D})
\end{equation}

\noindent
where $c$ ranges over all classes. This is not analytically tractable for deep NNs. Thus we use equation (\ref{eq9}) to approximate it as:

\vspace{-5mm}

\begin{equation}
    \hat{H}(\mathbf{\mathbf{y^*| \mathbf{x}^*, \mathcal{D}}}) = - \sum_c p_{\text{\tiny{MC}}}(\mathbf{y}^*=c|\mathbf{x}^*) \, \text{log} \, p_{\text{\tiny{MC}}}(\mathbf{y}^*=c|\mathbf{x}^*)
\end{equation}

\noindent
where $p_{\text{\tiny{MC}}}(\mathbf{y}^*=c|\mathbf{x}^*)$ is the average of the softmax probabilities of input $\mathbf{x}^*$ being in class $c$ over $T$ Monte Carlo samples. Finally, MI can be re-written as:

\vspace{-5mm}

\begin{equation}
\begin{aligned}
   & \hat{I}(\mathbf{\mathbf{y^*, w| \mathbf{x}^*, \mathcal{D}}}) = \hat{H}(\mathbf{y}^*|\mathbf{x}^*, \mathcal{D}) \\& + \sum_c \frac{1}{T} \sum_{t=1}^T  p(\mathbf{y^*}=c|\mathbf{x}^*, \hat{\mathbf{w}}_t) \, \text{log} \, p(\mathbf{y^*}=c|\mathbf{x}^*, \hat{\mathbf{w}}_t)
\end{aligned}
\end{equation}
\noindent
which can be computed for each model configuration at $t^{th}$ Monte Carlo run, $\hat{\mathbf{w}}_t$, obtained by the DropConnect.

\subsection{Unceratinty Evaluation Metrics}

MC-DropConnect proposed in the previous sections is a light-weight, scalable method to approximate Bayesian inference. This enables us to perform inference and estimate the uncertainty in deep neural networks at once. However, unlike model predictions, there is no ground truth for uncertainty values which makes evaluating the uncertainty estimates a challenging task. Therefore, there is no clear, direct approach to define a good uncertainty estimate.

\begin{figure*}[!t]
\begin{center}
\includegraphics[width=1.0\linewidth]{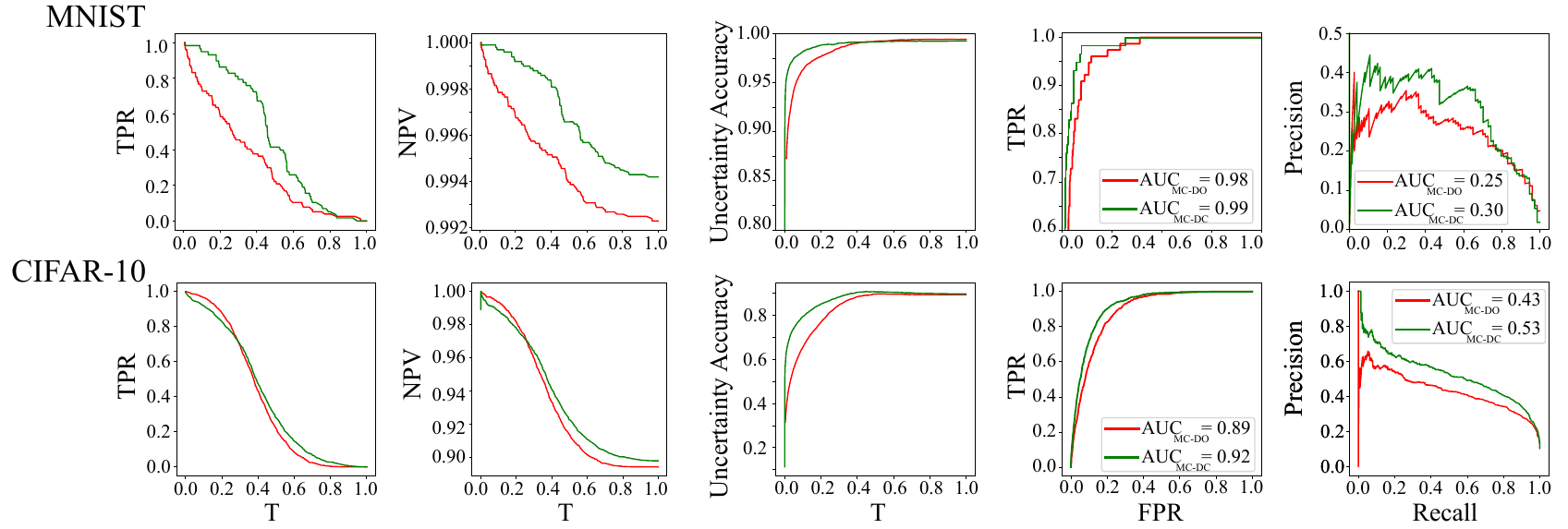}
\vspace{-5mm}
\end{center}
   \caption{\textbf{Illustrating the quantitative uncertainty estimation performance for the classification task using the proposed evaluation metrics.} Note that when varying the uncertainty threshold, our proposed MC-DropConnect approximated BNN  (shown in green) generally performs better than MC-Dropout  (shown in red) for both MNIST (Top) and CIFAR-10 (Bottom) datasets.}
\label{fig:mnist}
\end{figure*}

We propose metrics that incorporate the ground truth label, model prediction, and uncertainty value to evaluate the uncertainty estimation performance of such models. Fig. \ref{fig:uncertainty_metrics} shows the required processing steps to prepare these quantities for our metrics in a segmentation example. Note that these metrics can be used for both classification and semantic segmentation task as semantic segmentation is identical to the pixel-wise classification. Therefore, the conversions applied to a single pixel explains the classification task.

We first compute the map of \textit{correct} and \textit{incorrect} values by matching the ground truth labels and model predictions. Next, we convert the uncertainty map to a map of \textit{certain} and \textit{uncertain} predictions. This is done by setting an uncertainty threshold, $T_I$, which is changed between the minimum and maximum uncertainty values over the whole test set. Then the resulted correct/incorrect and certain/uncertain maps are treated as the new ground truth and predictions of a binary prediction task respectively. The following metrics can reflect the characteristics of a good uncertainty estimator:

\vspace{2mm}
\textbf{1. Negative predictive value (NPV):} If a model is certain about its prediction, the prediction should be correct. This can be written as a conditional probability:

\begin{equation}
P(\text{correct}|\text{certain}) = \frac{P(\text{correct}, \text{certain})}{P(\text{certain})} = \frac{\text{TN}}{\text{TN+FN}}
\end{equation}

\noindent
which is identical to the negative predictive value (NPV) measure used in a binary test.

\vspace{2mm}
\textbf{2. True Positive Rate (TPR):} If a model is making an incorrect prediction, it is desirable for the uncertainty to be high. 
\begin{equation}
P(\text{uncertain}|\text{incorrect}) = \frac{P(\text{uncertain}, \text{incorrect})}{P(\text{incorrect})} = \frac{\text{TP}}{\text{TP+FN}}
\end{equation}

\noindent
which is equivalent to the true positive rate (TPR) or recall measure commonly used in a binary test. In this scenario, the model is capable of flagging a wrong prediction with a high epistemic uncertainty value to help the user take further precautions.  

Note that the converse of the above two assumptions is not necessarily the case. This means that if a model is making a correct prediction on a sample, it does not necessarily require to be certain on the same. A model might, for instance, be able to correctly detect an object, but with a relatively higher uncertainty because it has rarely saw that instance in such pose or condition.

\vspace{2mm}
\textbf{3. Uncertainty Accuracy (UA):}
Finally, the overall accuracy of the uncertainty estimation can be measured as the ratio of the desired cases explained above (TP and TN) over all possible cases: 

\begin{equation}
\text{UA} = \frac{\text{TP}+\text{TN}}{\text{TP+TN+FP+FN}}
\end{equation}
 Clearly, for all the metrics proposed above, higher values correspond to the model that performs better. The value of these metrics depend on the uncertainty threshold, thus we plot each metric w.r.t the normalized threshold ($T$) and compare them using the area under each curve (AUC). This helps to summarize the value of each metric over various uncertainty thresholds in a single scalar. Note that in all the experiments, the normalized uncertainty threshold is computed using $T = \frac{u_T - u_{min}}{u_{max}-u_{min}}$ so that $T \in [0, 1]$ where $u_{min}$ and $u_{max}$ are the minimum and maximum uncertainty values over the whole dataset.

\section{Experimental Results and Discussion}
\label{results}
In this section, we assess the performance of uncertainty estimates obtained from DropConnect CNNs on the tasks of classification and semantic segmentation. We also compare the uncertainty obtained through our proposed method with the state-of-the-art method, MC-Dropout, on a range of datasets and show considerable improvement in prediction accuracy and uncertainty estimation quality. We quantitatively evaluate the uncertainty estimates using our proposed evaluation metrics. All the experiments are done using the TensorFlow framework \cite{abadi2016tensorflow}.

%-------------------------------------------------------------------------
\subsection{Classification}

\begin{figure}[!b]
\begin{center}
\includegraphics[width=1.0\linewidth]{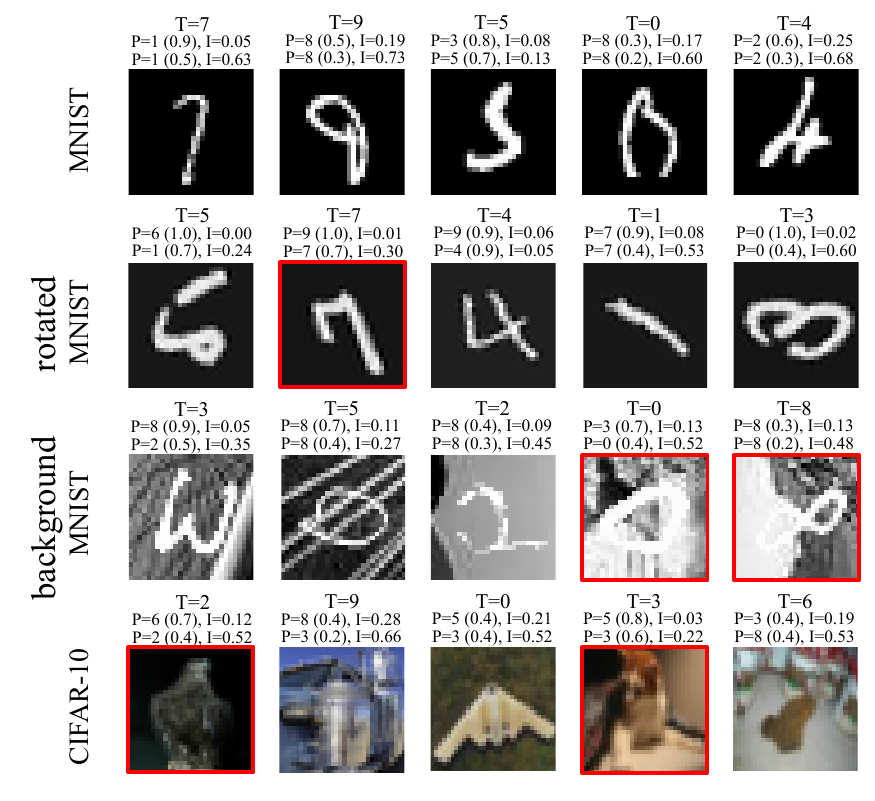}
\end{center}
   \caption{\textbf{Sample model prediction and uncertainty estimation results on MNIST, rotated MNIST, background MNIST, and CIFAR-10 datasets.} T: ground truth label, P: model prediction (with the average MC prediction probability of the predicted class provided in the parentheses), and I: model uncertainty estimation. For each sample, the second and third line of the provided information corressponds to MC-Dropout and MC-DropConnect respectively. The red boundary around images highlights FPs detected in MC-DropConnect prediction.}
\label{fig:classification_sample}
\end{figure}

We implement fully Bernoulli Bayesian CNNs using DropConnect to assess the theoretical insights explained above in the classification setting. We show that applying the mathematically principled DropConnect to all the weights of a CNN results in a test accuracy comparable with the state-of-the-art techniques in the literature while considerably improving the models' uncertainty estimation.

We picked LeNet structure (described in \cite{lecun1998gradient}) for the MNIST and a fully-convolutional network (FCNet) for the CIFAR-10 dataset. FCNet is composed of three blocks, each containing two convolutional layers (filter size of 3 and stride of 1) followed by a max-pooling (with filter size and stride of 2). The numbers of filters in the convolution layers of the three blocks are 32, 64, and 128, respectively. Each convolutional layer is also followed by a batch normalization layer and Relu non-linearity. We refer to the tests applied to the Bayesian CNN with DropConnect applied to all the weights of the network as ``MC-DropConnect'' and will compare it with ``None'' (with no dropout or dropconnect), as well as ``MC-Dropout'' \cite{gal2015bayesian} which has dropout used after all layers. 
To make the comparison fair, Dropout and DropConnect are applied with the same rate of $p=0.5$. We evaluate the networks using two testing techniques. The first is the standard test applied to each structure keeping everything in place (no weight or unit drop). The second test incorporates the Bayesian methodology the MC test equivalent to model averaging over $T=100$ stochastic forward passes. Note that the purpose of this experiment is not to achieve state-of-the-art results on either datasets, but to compare different models with different testing techniques.

\begin{figure}[!b]
\begin{center}
\includegraphics[width=1.0\linewidth]{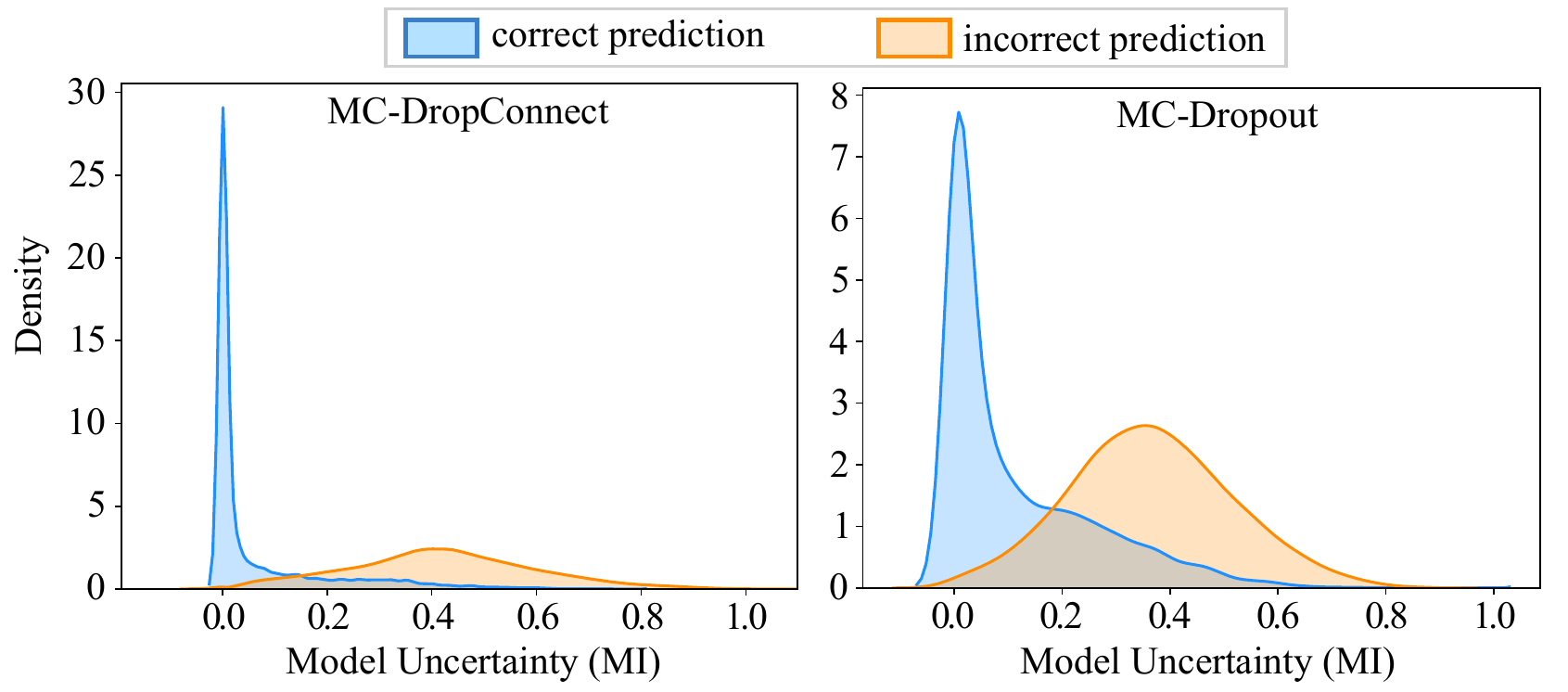}
\end{center}
   \caption{\textbf{Illustrating the distribution of model uncertainty values for the CIFAR-10 test samples.} Distributions are plotted separately for correct and incorrect predictions and for both MC-DropConnect (Left) and MC-Dropout (Right).}
\label{fig:distribution}
\end{figure}

Our experimental results (Table \ref{classification_result} and Fig. \ref{fig:mnist}) show that MC-DropConnect yields marginally improved prediction accuracy when applying MC-sampling. More importantly, the uncertainty estimation metrics show a significant improvement when using MC-DropConnect. Example predictions are provided in Fig. \ref{fig:classification_sample}. We also test the LeNet networks (trained on MNIST) on rotated and background MNIST data. These are the distorted versions of MNIST which can be counted as the out-of-distribution examples \cite{hendrycks2016baseline} that the model has never seen before. This is done to check the quality of predictive uncertainty in terms of its generalization to domain shift.

As shown in Fig. \ref{fig:classification_sample}, MC-DropConnect BNN often yields high uncertainty estimate when the prediction is wrong and makes accurate predictions when it is certain. We observed fewer failure cases using MC-DropConnect compared with MC-Dropout (also reflected in the TPR and NPV depicted in Fig. \ref{fig:mnist}). This is also indicated in Fig. \ref{fig:distribution} which shows the distribution of the model uncertainty over the correct and incorrect predictions separately. It shows that the MC-DropConnect approximation produces significantly higher model uncertainty values (Kolmogorov-Smirnov test yields p-value $<0.001$) when the prediction is erroneous. Thus, this adds complementary information to the conventional network output which can be leveraged by the automated system to reject the prediction and send it for further inspection. 

We also observed interesting, meaningful FPs (correct predictions with high uncertainty estimation) when using MC-DropConnect. Examples are highlighted with red boundaries in Fig. \ref{fig:classification_sample}. These FPs are often corresponding to the visually more complicated samples where the network is not confident about. Such FPs are pleasant and considered as red flags that a model is more likely to make inaccurate predictions.

\begin{figure}[!t]
\begin{center}
\includegraphics[width=1.0\linewidth]{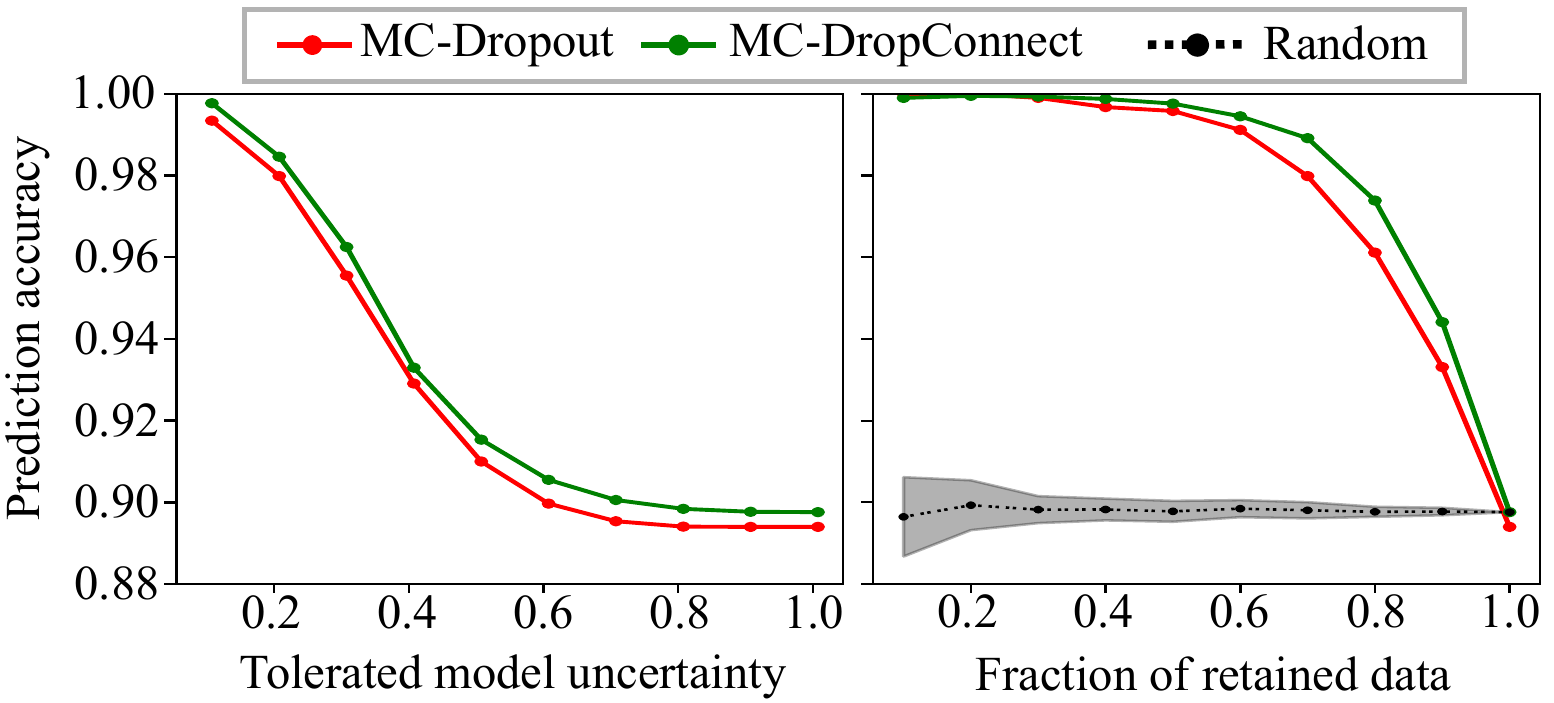}
\end{center}
   \caption{\textbf{Enhanced prediction accuracy achieved via rejecting the highly uncertain samples.} The prediction accuracy is computed over the test samples of the CIFAR-10 dataset and depicted as a function of the tolerated amount of model uncertainty (Left), and retained data size. The black curve in the right panel illustrates the effect of randomly rejecting the same number of samples. It is plotted as mean ($\pm$std) over 20 sampling. This shows uncertainty is an effective measure of prediction accuracy.}
\label{fig:acc_frac}
\vspace{-4mm}
\end{figure}

\vspace{1mm}
\textbf{Enhanced performance with uncertainty-informed referrals:}
An uncertainty estimation with such characteristics (i.e. high uncertainty as an indication of erroneous prediction, reasonable FPs) provides valuable information in situations where the control is handed to automated systems in real-life settings, with the possibility of becoming life-threatening to humans. These include applications such as self-driving cars, autonomous control of drones, automated decision making and recommendation systems in the medical domain, etc.

\begin{figure}[!t]
\begin{center}
\includegraphics[width=0.9\linewidth]{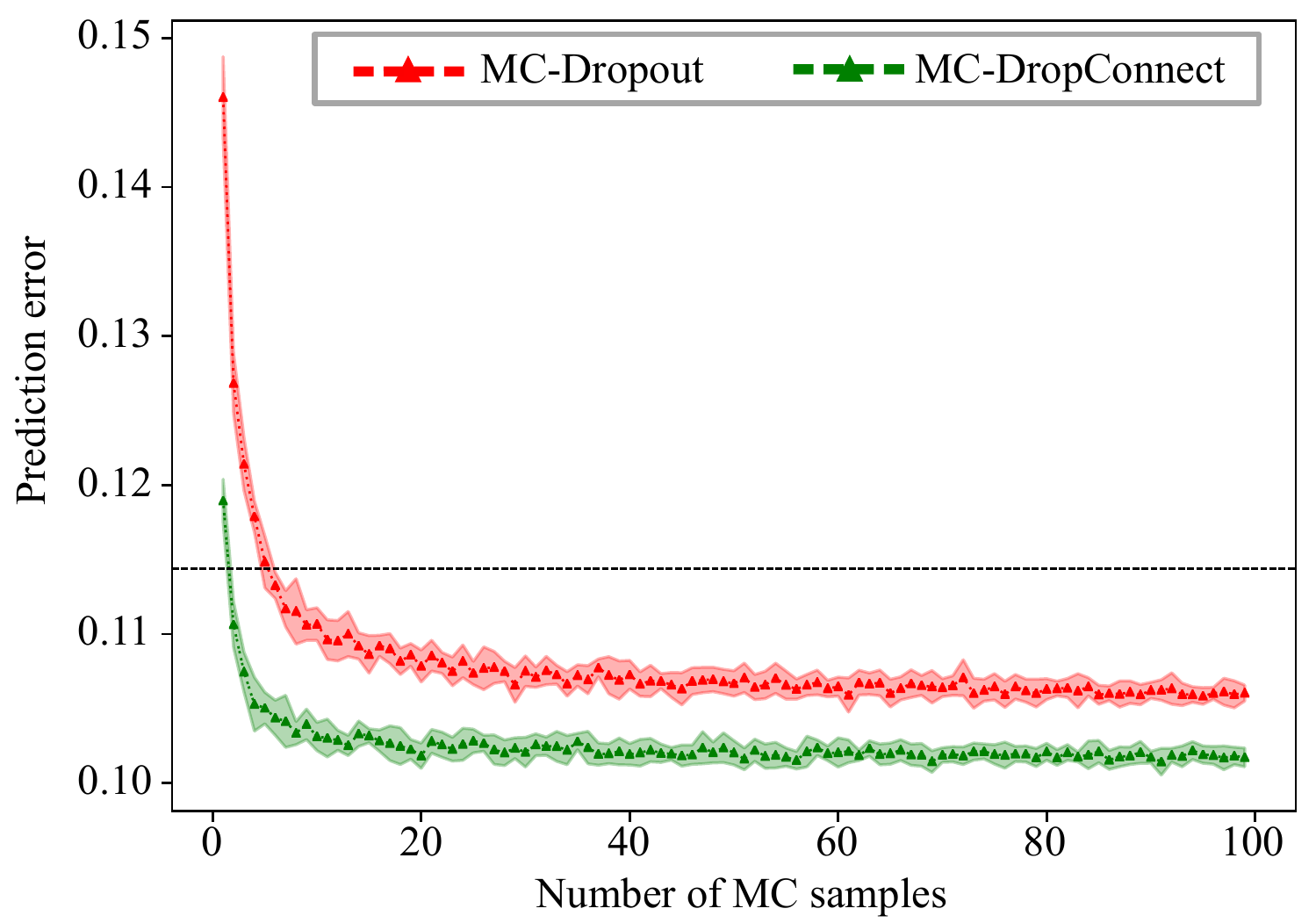}
\end{center}
   \caption{\textbf{Test error of the FCNet on CIFAR-10 for different numbers of forward-passes in MC-Dropout and MC-DropConnect, averaged with 10 repetitions.} The shaded area around each curve shows one standard deviation. The black dotted line shows the test error for the same neural network with no sampling.}
\label{fig:convergence}
\vspace{-5mm}
\end{figure}

\begin{table*}[!t]
\centering
\caption{Quantitative prediction and uncertainty estimation performance of the various frameworks on the CamVid, CityScapes, and CT-Organ datasets. Our quantitative analyses supports the superior performance of the MC-DropConnect in terms of both segmentation accuracy and uncertainty estimation quality.}
\begin{tabular}{|l|l|ccc|ccc|}
\hline
\multicolumn{1}{|c|}{\multirow{2}{*}{Data (Model)}} & \multicolumn{1}{c|}{\multirow{2}{*}{\begin{tabular}[c]{@{}c@{}}Uncertainty \\ Estimation Method\end{tabular}}} & \multicolumn{3}{c|}{Prediction Performance (\%)} & \multicolumn{3}{c|}{Uncertainty metrics AUC (\%)} \\ \cline{3-8} 
\multicolumn{1}{|c|}{} & \multicolumn{1}{c|}{} & Pixel accuracy & Mean accuracy & Mean IOU & TPR & NPV & UA \\ \hline
\multirow{3}{*}{CamVid (SegNet)} & None & 79.46 & 65.03 & 46.31 & - & - & - \\
 & MC-Dropout & 80.99 & 65.46 & 47.31 & 17.23 & 82.48 & 80.18 \\
 & MC-DropConnect & \textbf{82.92} & \textbf{67.47} & \textbf{49.53} & \textbf{21.63} & \textbf{86.54} & \textbf{82.78} \\ \hline
\multirow{3}{*}{CityScapes (ENet)} & None & 87.50 & 55.30 & 44.08 & - & - & - \\
 & MC-Dropout & 87.38 & 56.35 & 44.11 & 6.12 & 88.67 & 84.89 \\
 & MC-DropConnect & \textbf{88.87} & \textbf{63.83} & \textbf{50.25} & \textbf{9.61} & \textbf{90.33} & \textbf{85.57} \\ \hline
 \multirow{3}{*}{CT-Organ (VNet)} & None & 95.19 & 96.44 & 65.49 & - & - & - \\
& MC-Dropout & 94.11 & \textbf{97.73} & 67.07 & \textbf{10.81} & 86.41 & 91.51 \\
 & MC-DropConnect & \textbf{97.90} & 97.71 & \textbf{72.77} & 6.69 & \textbf{87.03} & \textbf{92.59} \\ \hline
\end{tabular}
\label{seg_table}

\end{table*}

\begin{figure*}[!b]
\begin{center}
\includegraphics[width=1.0\linewidth]{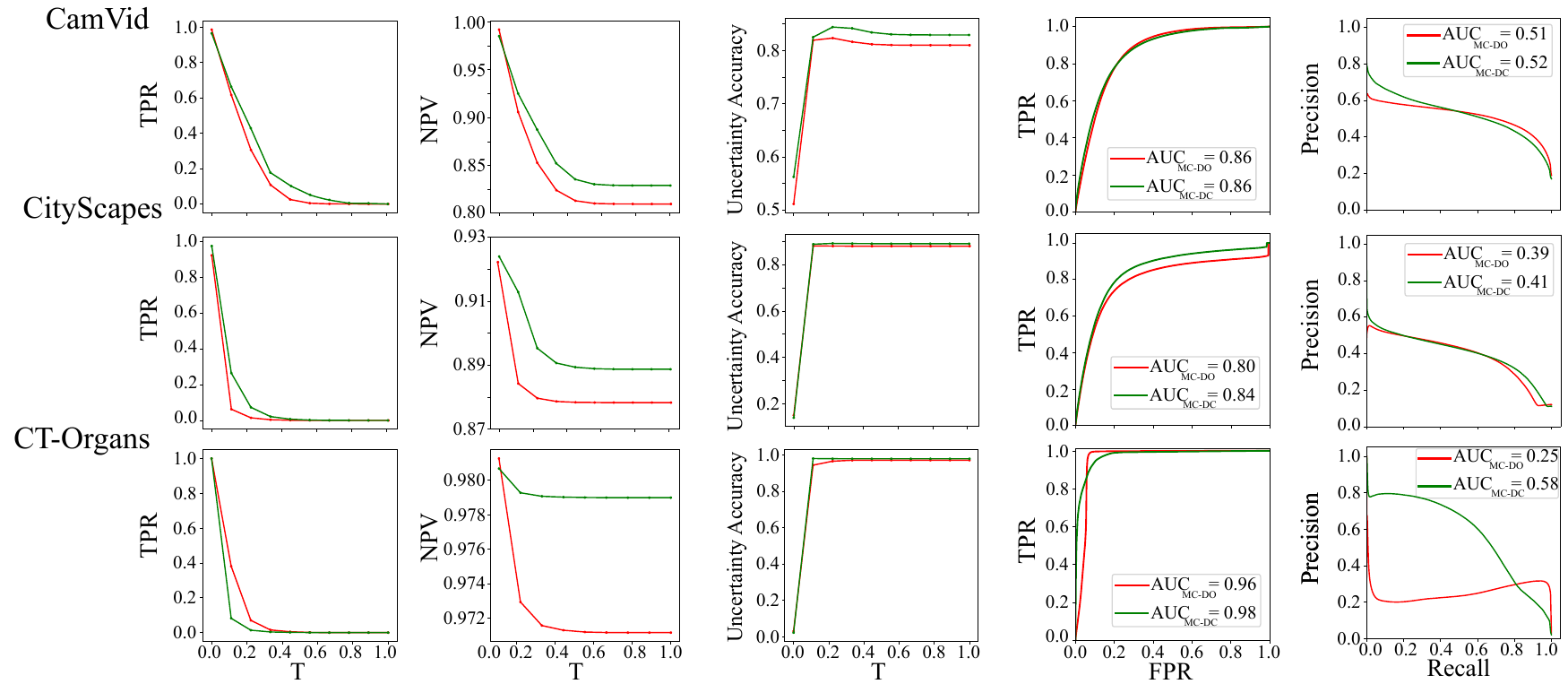}
\end{center}
   \caption{\textbf{Illustrating the quantitative uncertainty estimation performance for the semantic segmentation task using the proposed evaluation metrics.} Note that when varying the uncertainty threshold, our proposed MC-DropConnect approximated BNN (shown in green) generally performs better than MC-Dropout (shown in red) for CamVid (Top) and CityScapes (Middle), and CT-Organs (Bottom) datasets.}
\label{fig:seg_result}
\end{figure*}

An automated cancer detection system, for example, trained on a limited number of data (which is often the case due to the expensive or time-consuming data collection process) could encounter test samples lying out of its observed data distribution. Therefore, it is prone to make unreasonable decisions or recommendations which could finally result in experts' bias. However, uncertainty estimation can be utilized in such scenarios to detect such undesirable behavior of the automated systems and enhance the overall performance by flagging appropriate subsets for further analysis. 

We set up an experiment to test the usefulness of the proposed uncertainty estimation in mimicking the clinical work-flow, and referring samples with high uncertainty for further testing. First, the model predictions are sorted according to their corresponding epistemic uncertainty (measured by the mutual information metric). We then computed the prediction accuracy as a function of confidence. This is done by taking various levels of tolerated uncertainty and the fraction of retained data (see Fig. \ref{fig:acc_frac}). We observed a monotonic increase in prediction accuracy with MC-DropConnect outperforming MC-Dropout for decreasing levels of tolerated uncertainty and decreasing fraction of retained data. It is also compared with removing the same fraction of samples randomly, that is with no use of uncertainty information, which indicates the informativeness of the uncertainty about prediction performance as well.

\vspace{2mm}
\textbf{Convergence of the MC-DropConnect}
Even though the proposed MC-DropConnect method results in better prediction accuracy and uncertainty estimation, it still comes with a price of prolonged test time. This is because we need to evaluate the network stochastically multiple times and average the results. Therefore, while the training time of the models and their probabilistic variant is identical, the test time is scaled by the number of averaged forward passes. This becomes more important in practice and for applications which the test-time efficiency is critical. To evaluate the MC-DropConnect approximation method, we assessed the prediction accuracy of the FCNet on CIFAR-10 dataset and for different number of Monte Carlo simulations (T). We then reported the average results over 10 runs in Fig. \ref{fig:convergence}. We see that MC-DropConnect results in a significantly lower prediction error than the baseline network after only 2 samples while this number is 6 for MC-Dropout. Moreover, MC-Dropconnect achieves an error less than one standard deviation away from its best performance (at T=90) after only 18 samples, while this number is 54 for MC-Dropout (with its best performance at T=94).

\begin{figure*}[!t]
\begin{center}
\includegraphics[height=0.99\linewidth]{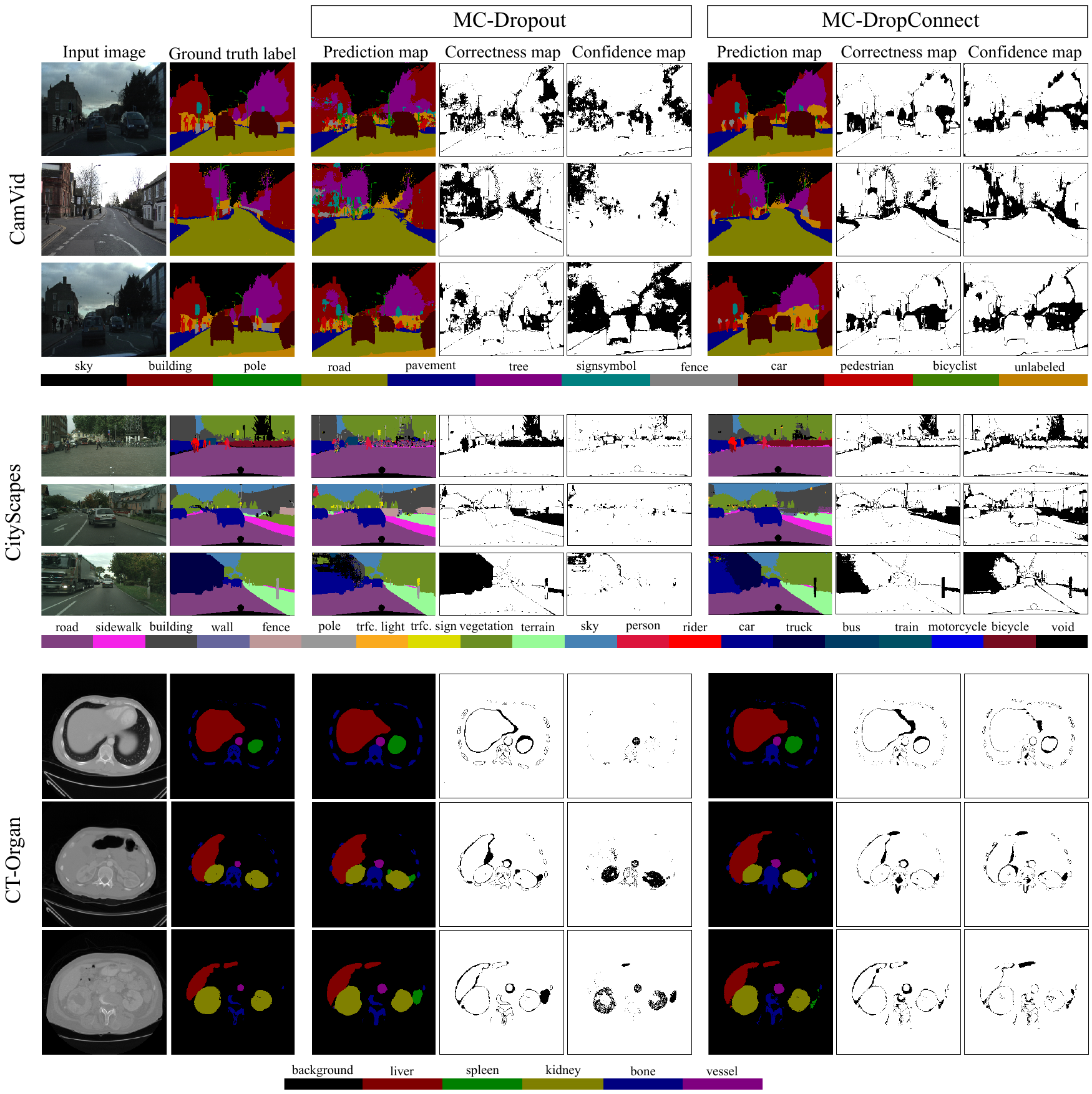}

\end{center}
   \caption{\textbf{Qualititative results for semantic segmentation and uncertainty estimates on CamVid, CityScapes, and CT-Organs datasets.} Each row depicts a single sample and includes the input image with ground truth, prediction, correctness, and confidence (using the mutual information metric) maps for both MC-Dropout and MC-DropConnect. Correctness map is the binary map that shows the correct and incorrect predictions. Confidence map is the thresholded map of uncertainty values computed over all classes. In all cases, the threshold is set manually to the one that achieves the highest UA. Correct and certain regions are respectively shown in white color in correctness and confidence maps.}
\label{fig:seg_samples}
\end{figure*}

\subsection{Semantic Segmentation}
In this section, we perform similar experiments to assess the performance of MC-DropConnect approximation of the BNNs and compare it with the benchmark MC-Dropout. The segmentation prediction performance is quantified using the pixel accuracy, mean accuracy and mean IOU metrics defined in \cite{long2015fully}. Details of the datasets and network architectures used in each of the experiments are explained below briefly. Note that in all the experiments, dropout and dropconnect layers are placed in the same part of the network and with the same rate of $p=0.5$.

\textbf{CamVid with SegNet:}
CamVid \cite{brostow2009semantic} is a road scene understanding dataset which contains 367, 100, and 233 training, validation, and test images respectively, with 12 classes. Images are size $360 \times 480$ and include both bright and dark scenes. We chose SegNet as the network architecture to be used for the semantic segmentation task to make the results of our approach to those of \cite{kendall2015bayesian}.

\textbf{CityScapes with ENet:}
CityScapes is one of the most popular datasets for the urban scene understanding with 5000, 500, and 1525 images for training, validation, and test. Images are of size $2048 \times 1024$ collected in 50 different cities and contains 20 different classes. Due to the large size of the images and more number of classes, we chose ENet \cite{paszke2016enet} which is a more powerful network that requires much less flop and parameters. The spatial dropout layers used in this framework is replaced with the regular dropout and dropconnect layers for our purpose.

\textbf{3D CT-Organ with VNet:}
Since uncertainty estimates can play a crucial role in the medical diagnostics field, we also tested our model uncertainty estimation approach in the semantic segmentation of the body organs in abdominal 3D CT-scans. We used V-Net \cite{milletari2016v} which is one of the commonly used architectures for the segmentation of the volumetric medical images. The data include six classes including background, liver, spleen, kidney, bone, and vessel.

\textbf{Qualititative observations}. Fig \ref{fig:seg_samples} shows example segmentation and model uncertainty results from the various Bayesian frameworks on different datasets. This figure also compares the qualitative performance of MC-DropConnect with that of MC-Dropout. The correctness and confidence map highlights the misclassified and uncertain pixels respectively. Our observations show that MC-Dropconnect produces high-quality uncertainty estimation maps outperforming MC-Dropout, i.e. displays higher model uncertainty when models make wrong predictions.

We generally observe that higher uncertainty values associate with three main scenarios. First, at the boundaries of the object classes (capturing the ambiguity in labels transition). Second, we observe a strong relationship between the frequency at which a class label appears and the model uncertainty. Models generally have significantly higher uncertainty for the rare class labels (the ones that are less frequent in the data; such as pole and sign symbol classes in CamVid). Conversely, models are more confident about class labels that are more prevalent in the datasets. Third, models are less confident in their prediction for objects that are visually difficult or ambiguous to the model. For example, (bicyclist, pedestrian) classes in CamVid and (car, truck) classes in CityScapes are visually similar which makes it difficult for the model to make a correct prediction, thus displays higher uncertainty.

\textbf{Quantitative observations}. We report the semantic segmentation results in Table \ref{seg_table} and Fig. \ref{fig:seg_result}. We find that MC-DropConnect generally improves the accuracy of the predicted segmentation masks for all three model-dataset pairs. 

Similar to what is done in the classification task, we computed the segmentation accuracies for varying levels of model confidence. The results are provided in Table \ref{confidence}. For all three dataset-model pairs, we observed very high levels of accuracy for the 90th percentile confidence. This indicates that the proposed method results in the model uncertainty estimate which is an effective measure of confidence in the prediction.

\begin{table}[]
\centering
\caption{Pixel-wise accuracy of the Bayesian frameworks as a function of confidence for the 0th percentile (all pixels) through to the 90th percentile (10\% most certain pixels).  This shows uncertainty is an effective measure of prediction accuracy.}
\begin{tabular}{c|ccc}
& \multicolumn{3}{c}{Pixel-wise classification accuracy} \\
\begin{tabular}[c]{@{}c@{}}Confidence\\ percentile\end{tabular} & CamVid & CityScapes & CT-Organ  \\ \hline \hline
0 & 82.92 & 88.87  & 97.90  \\
10 & 87.45 & 90.59 & 99.81  \\
50 & 97.83 & 92.13 & 99.97   \\
90 & 93.68 & 99.32 & 99.99           
\end{tabular}
\label{confidence}
\end{table}

\section{Conclusion}
\label{conclusion}
We have presented MC-DropConnect as a mathematically grounded and computationally tractable approximate inference in Bayesian neural networks. This framework outputs a measure of model uncertainty with no additional computational cost; i.e. by extracting the information from the existing models that have been thrown away so far. We also developed new metrics to evaluate the uncertainty estimation of the models in all ML tasks, such as regression, classification, semantic segmentation, etc. 
We created the probabilistic variants of some of the most famous frameworks (in both classification and semantic segmentation tasks) using MC-DropConnect. Then we exploited the proposed metrics to evaluate and compare the uncertainty estimation performance of various models.

Empirically, we observed that the MC-DropConnect improves the prediction accuracy, and yields a precise estimation of the model confidence to its prediction. Analysis of the output uncertainty estimate via the proposed metrics shows that the model uncertainty estimates serve as an additive piece of information which can assist users in the decision-making process.

Future research includes the study of how imposing the dropconnect (and with different drop probabilities) affects the trained convolutional kernels. Leveraging the uncertainty in the training process to enrich the model's knowledge of the data domain is another interesting research direction.

%\section*{Acknowledgment}


\ifCLASSOPTIONcaptionsoff
  \newpage
\fi



%\bibliographystyle{IEEEtran}

% \end{thebibliography}
\bibliographystyle{IEEEtran}
\bibliography{references}



%\vfill


\end{document}